\documentclass[10pt,twocolumn,letterpaper]{article}

\usepackage[pagenumbers]{cvpr} % To force page numbers, \eg for an arXiv version
\usepackage[accsupp]{axessibility}  % Improves PDF readability for those with disabilities.

\definecolor{cvprblue}{rgb}{0.21,0.49,0.74}
\usepackage[pagebackref,breaklinks,colorlinks,allcolors=cvprblue]{hyperref}
\usepackage{multirow}
\usepackage{diagbox}
\usepackage[misc]{ifsym}
\usepackage{amssymb}
\usepackage{caption}
\usepackage{threeparttable}
\usepackage{colortbl}
\usepackage{xcolor}
\usepackage{algorithmic,algorithm}
\usepackage[accsupp]{axessibility}
\usepackage{graphicx}
\usepackage{rotating}
\def\paperID{xxxxx} % *** Enter the Paper ID here
\def\confName{CVPR}
\def\confYear{2026}
\newcommand\blfootnote[1]{%
  \begingroup
  \renewcommand\thefootnote{}\footnote{#1}%
  \addtocounter{footnote}{-1}%
  \endgroup
}

\begin{document}
%%%%%%%%% TITLE - PLEASE UPDATE
\title{Virtual Full-stack Scanning of Brain MRI via Imputing Any Quantised Code}

%%%%%%%%% AUTHORS - PLEASE UPDATE
\author{Yicheng Wu\textsuperscript{1,2 (\Letter, *)}, Tao Song$^{3,*}$,
Zhonghua Wu$^{4}$, 
Jin Ye$^{2}$,
Zongyuan Ge$^{2}$,
Wenjia Bai$^{1}$, \\
Zhaolin Chen$^{2}$,
and Jianfei Cai$^{2}$
\\
$^{1}$Imperial College London, $^{2}$Monash University, \\
$^{3}$Fudan University, $^{4}$Nanyang Technological University\\
}

\maketitle

\blfootnote{$^*$ Equal Contribution. Correspondence to y.wu2@imperial.ac.uk.}

\maketitle
\begin{abstract}
Magnetic resonance imaging (MRI) is a powerful and versatile imaging technique, offering a wide spectrum of information about the anatomy by employing different acquisition modalities. However, in the clinical workflow, it is impractical to collect all relevant modalities due to the scan time and cost constraints.
Virtual full-stack scanning aims to impute missing MRI modalities from available but incomplete acquisitions, offering a cost-efficient solution to enhance data completeness and clinical usability. Existing imputation methods often depend on global conditioning or modality-specific designs, which limit their generalisability across patient cohorts and imaging protocols.
To address these limitations, we propose \textbf{CodeBrain}, a unified framework that reformulates various ``any-to-any'' imputation tasks as a region-level full-stack code prediction problem. CodeBrain adopts a two-stage pipeline: (1) it learns the compact representation of a complete MRI modality set by encoding it into scalar-quantised codes at the region level, enabling high-fidelity image reconstruction after decoding these codes along with modality-agnostic common features; (2) it trains a projection encoder to predict the full-stack code map from incomplete modalities via a grading-based design for diverse imputation scenarios.
Extensive experiments on two public brain MRI datasets, i.e., IXI and BraTS 2023, demonstrate that CodeBrain consistently outperforms state-of-the-art methods, establishing a new benchmark for unified brain MRI imputation and enabling virtual full-stack scanning.
Our code will be released at \url{https://github.com/ycwu1997/CodeBrain}.
\end{abstract}

\section{Introduction}
\label{sec:intro}
Magnetic resonance imaging (MRI) is widely used in clinical practice and research due to its excellent soft tissue contrast, providing essential information for disease diagnosis, prognosis, and longitudinal monitoring of outcomes \cite{lee2020assessing,gilmore2012longitudinal,wu2024dataset}.
For the brain, a typical MRI examination involves multiple acquisition protocols, each producing a distinct modality that emphasises specific anatomical or pathological characteristics. For example, T1-weighted (T1) scans delineate anatomical structures, fluid-attenuated inversion recovery (FLAIR) scans are sensitive to lesions, while gadolinium contrast-enhanced T1-weighted (T1Gd) scans highlight active tumour regions.
However, acquiring a complete set of MRI modalities is time-consuming and thus often impractical in routine clinical workflows. Certain modalities, such as T1Gd, require the injection of a contrast agent that adds extra waiting time and poses potential health risks \cite{wang2023quantitative}. Moreover, motion-induced artifacts and protocol variability may cause mis-registration and inconsistent image quality \cite{balakrishnan2019voxelmorph}. These limitations motivate the development of modality imputation methods that can synthesise missing scans and provide a complete modality set of brain MRI to enhance clinical applications  \cite{xie2023cross}.

\begin{figure}[t]
  \centering
   \includegraphics[width=1\linewidth]{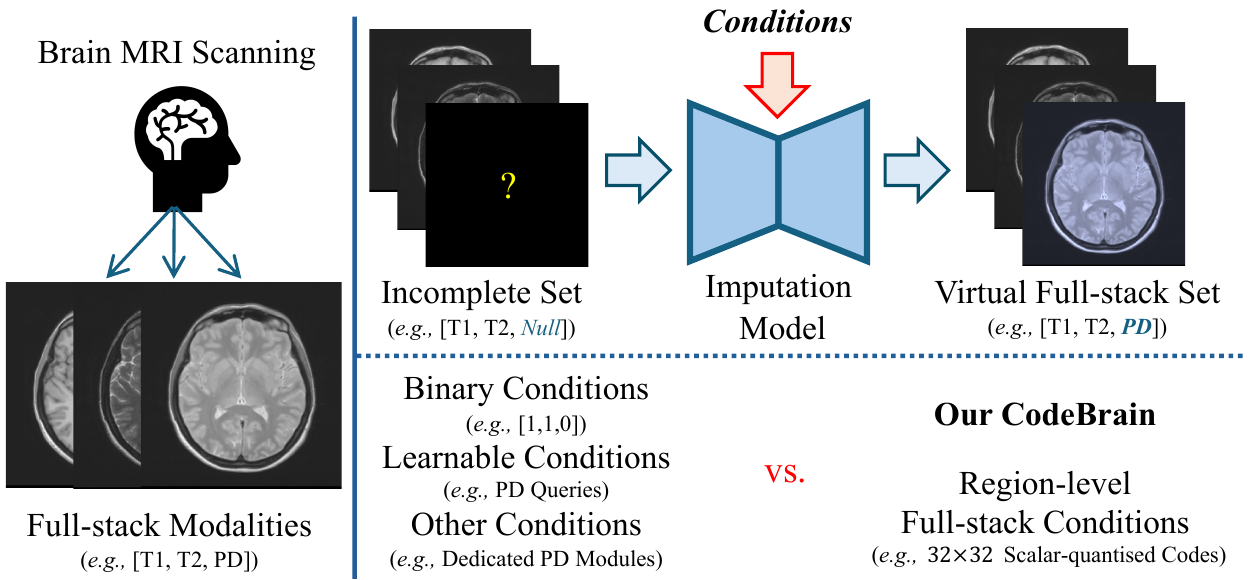}
   \caption{\textbf{Illustration of virtual full-stack scanning}. Given an incomplete brain MRI modality set (\eg, T1 and T2 only), the model imputes the missing modality (\eg, PD) according to the specified condition. Unlike existing approaches that rely on global indicators (\eg, binary vectors such as [1, 1, 0] or learnable modality queries) or dedicated modality-specific modules, \textbf{CodeBrain} employs region-level conditions, \eg, 32$\times$32 scalar-quantised codes that spatially represent the full-stack MRI, to provide fine-grained guidance for the unified imputation of brain MRI.}
   \label{fig_concept}
\end{figure}

Deep learning-based imputation models have shown promising progress in medical imaging \cite{dayarathna2023deep}. Theoretically, at each pixel, the intensities of two different MRI modalities for the same subject share the same underlying spin properties \cite{zhao2025reverse}, and thus are transferable. Practically, SynthSeg \cite{billot2023synthseg} achieves robust brain structure segmentation across diverse modalities, suggesting shared structural priors in images of different modalities. Due to transferable and shared information, missing MRI modalities could possibly be inferred from existing available ones.
To this end, various technical designs have been explored for modality imputation, including generative adversarial networks (GANs) \cite{sharma2019missing,dar2019image}, transformer architectures \cite{dalmaz2022resvit,liu2023one}, and recently diffusion models \cite{meng2024multi,arslan2025self}. 
As illustrated in Fig.~\ref{fig_concept}, most existing methods depend on global conditions, such as binary task indicators \cite{meng2024multi} or learnable modality prompts \cite{liu2023one}, to specify available and missing modalities. However, global conditions may fail to capture region-level and cross-modality variability inherent to MRI data. To address this, some works introduce modality-specific modules or decoders \cite{liu2023one,zhang2024unified}. But these designs substantially increase the model parameters and scale poorly when more modalities are involved, thus limiting generalisability across datasets.

These challenges motivate us to develop \textbf{CodeBrain}, a unified framework for modality imputation to enable virtual full-stack scanning of brain MRI.
The key idea is to reformulate the complex ``\textit{any-to-any}'' imputation problem into a simpler region-level code prediction task.
Instead of synthesising each missing modality independently, CodeBrain predicts a compact representation of the entire modality set that encapsulates structural and intensity relationships across modalities.
To further enhance spatial adaptivity, these cross-modality relationships are represented as scalar-quantised codes at the region level.
This design preserves fine-grained variations and enables localised imputation that global conditioning approaches fail to achieve.

Specifically, CodeBrain adopts a two-stage pipeline.
In the first stage, a reconstruction model learns to disentangle a complete brain MRI modality set into two components:
(1) shared, modality-agnostic features capturing the anatomical structure;
(2) region-level scalar-quantised codes, named as full-stack codes, that compactly represent modality-specific information.
The complete MRI set can be reconstructed by decoding the full-stack codes together with shared features.
In the second stage, a projection encoder is trained to predict these full-stack codes from incomplete modalities, guided by a grading-based objective, so that imputation can be performed using the predicted full-stack codes.

Our main contributions are summarised as follows:
\begin{itemize}
\item We introduce a novel paradigm for virtual full-stack scanning of brain MRI that reformulates the challenging \textit{Any-to-Any} imputation task into a unified full-stack code prediction problem, enabling direct modelling of cross-modality relationships and subject-level characteristics.

\item We design CodeBrain as a two-stage training framework. The first stage extracts modality-agnostic features and learns specific quantized codes through a reconstruction process. The second stage predicts these full-stack codes by using a customized grading loss. 

\item Extensive experiments demonstrate that CodeBrain outperforms five state-of-the-art methods, achieving superior imputation quality and setting a new benchmark for brain MRI modality imputation. In addition, we demonstrate that modalities imputed by CodeBrain improve the downstream clinical task for brain tumour segmentation.
\end{itemize}

\section{Related Work}
\label{sec:related_work}
\noindent\textbf{MRI Modality Imputation.}
Modality incompleteness is a prevalent challenge in various medical imaging applications. To address this, deep learning models have been widely applied to impute the missing data, in particular via cross-modality translation \cite{cao2023autoencoder,dayarathna2023deep,ozbey2023unsupervised,gungor2023adaptive}. For example, \cite{pan2018synthesizing} proposed a Cycle-GAN model to generate the corresponding PET from MRI to aid in Alzheimer’s disease diagnosis. An image-to-image translation model was proposed in \cite{wang2023quantitative} to synthesise the cerebral blood volume data from standard MRI modalities, enhancing its clinical applicability.

Recent studies have focused on unified MRI imputation \ie, using a single model to perform imputation for different modalities. For instance, MMGAN \cite{sharma2019missing} employed the GAN model to improve the synthesis quality. MMT \cite{liu2023one} utilised a Swin Transformer with modality-dependent queries to generate missing modalities. M2DN \cite{meng2024multi} employed a diffusion model with binary conditional codes to specify different imputation tasks. Furthermore, \cite{zhang2024unified} explored the use of modality-shared and modality-specific modules to improve unified imputation performance. \cite{dalmaz2024one} incorporated federated learning for multi-modal MRI synthesis. MMHVAE \cite{dorent2025unified} proposed a mixture-of-expert model in the variational auto-encoder architecture for modality translation.  \cite{tudosiu2024realistic} applied an auto-regressive generation pipeline for brain MRI synthesis, ensuring morphological consistency. \cite{zhang2025structure} improves latent diffusion using specific and shared encoders, and \cite{song2026learning} achieves adaptive interactions among different channels.

Most existing unified approaches perform pixel-to-pixel modality translation. In contrast, our CodeBrain performs cross-modality translation in the quantised latent space, predicting the code for a missing modality and synthesising its image with both the predicted code and the extracted common features. It offers a flexible framework and eliminates the need for designing modality-specific modules.

\begin{figure*}[htbp]
  \centering
   \includegraphics[width=1\linewidth]{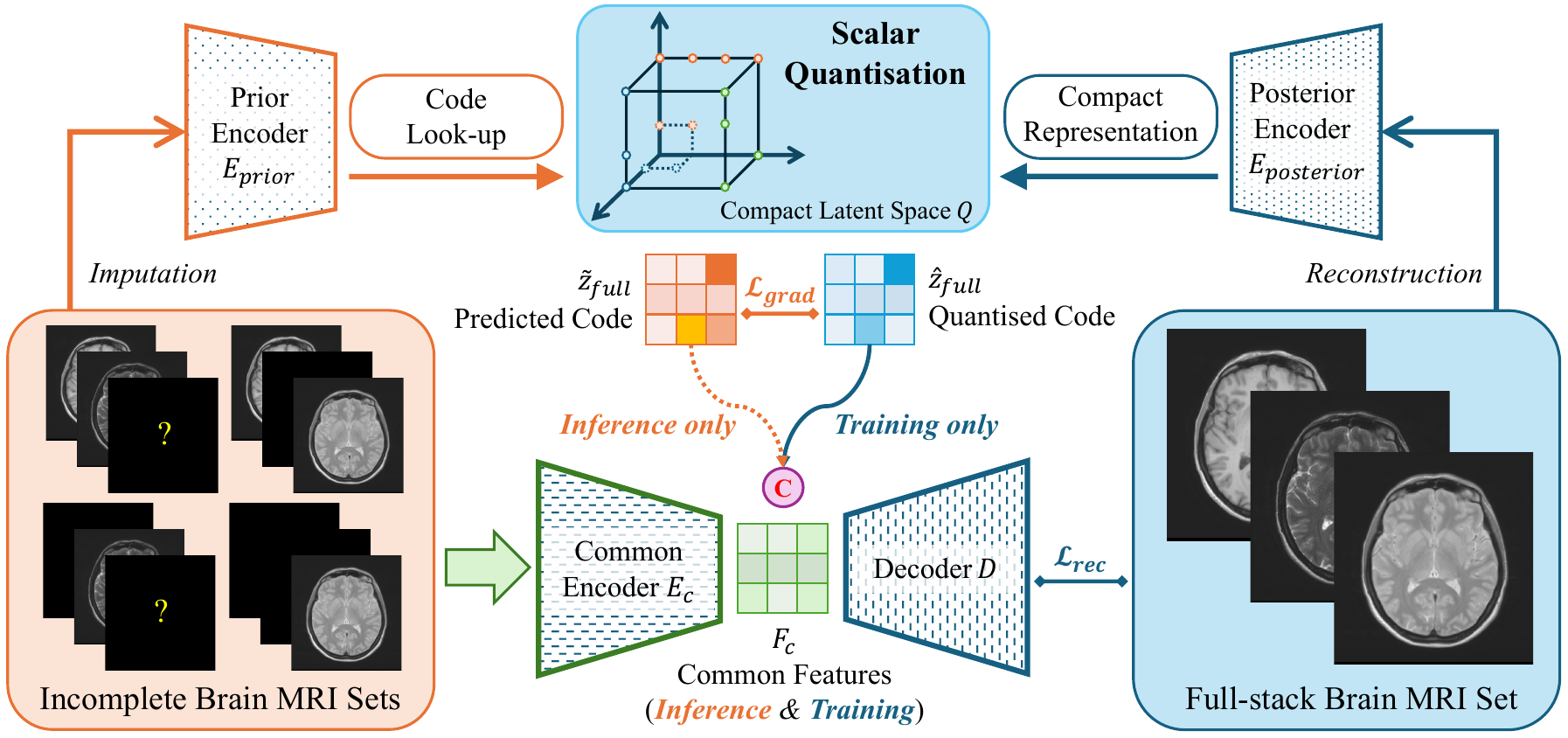}
   \caption{\textbf{Overall diagram of CodeBrain.} First, it constructs a compact latent space $Q$. A complete brain MRI set of all modalities is projected by $E_{posterior}$ to its quantised codes. With the code, images of all modalities can be reconstructed by a decoder $D$ using the code along with common features $F_c$, the latter being inferred from available modalities using $E_c$. Second, an encoder $E_{prior}$ is trained by a grading loss to predict these codes. During inference, these predicted codes serve as region-level conditions to guide the imputation.}
   \label{fig_pipeline}
\end{figure*}

\noindent\textbf{Quantisation-based Image Generation.}
Quantisation projects complex data into a discrete and low-dimensional representation, facilitating multi-modal alignment, mapping, and fusion \cite{he2021checkerboard,hsu2023disentanglement}. Vector quantisation (VQ) techniques have been applied in a range of tasks, including image reconstruction, transformation, and generation \cite{esser2021taming}. For example, VQVAE \cite{van2017neural} introduced the neural discrete representation, while VQGAN \cite{esser2021taming} employed a patch GAN model to capture fine details and a causal transformer to predict quantised codes for high-resolution image synthesis. Straight-through estimator \cite{van2017neural} and Gumbel-softmax \cite{jang2016categorical} were proposed to back-propagate gradients for the quantisation operator, enabling end-to-end model training. To improve the image generation quality, residual designs \cite{lee2022autoregressive}, multi-scale codebooks \cite{razavi2019generating,williams2020hierarchical}, dynamic token embeddings \cite{huang2023towards}, and codebook splitting \cite{zheng2022movq} have been proposed to improve VQ representation capacity. Recently, beyond region-level VQ, global quantisation methods \cite{tian2024visual,yu2024image,han2025infinity} have been explored to represent multiple image-level information. Furthermore, VQ has been incorporated with the diffusion and transformer models \cite{gu2022vector,rombach2022high,tang2024hart} to produce high-quality generation results. 

However, training a robust codebook is still a critical bottleneck for VQ-based models \cite{mentzerfinite,zhu2025addressing}. Therefore, implicit codebooks have been explored, where the decoder reconstructs the image from only code indices instead of using the nearest code vectors \cite{hao2024bigr}. For example, BSQ \cite{zhaoimage} used a sign function to achieve binary quantisation. FSQ \cite{mentzerfinite} proposed a straightforward rounding operation for quantisation, discarding the complex process of learning a codebook. Inspired by previous works, the proposed CodeBrain model generates codes in a finite scalar-quantised space as \cite{mentzerfinite}, which can be seen as the region-level compact representation of full-stack brain MRI.

\section{Method}
\label{sec:method}
Fig.~\ref{fig_pipeline} provides an overview of the proposed \textbf{CodeBrain} framework. The core idea is to reformulate cross-modality image translation as a cross-modality code prediction task, predicting quantised codes for the missing modality within a compact latent space $Q$. The image for the missing modality can then be recovered by decoding its corresponding codes together with modality-agnostic common features, the latter being extracted from available modalities.
Our training process consists of two stages:
Stage \uppercase\expandafter{\romannumeral1} constructs a finite scalar space and quantises each full-stack brain MRI into a compact code map, while Stage \uppercase\expandafter{\romannumeral2} learns to predict these codes from the available modalities, supervised by the corresponding code maps generated in Stage \uppercase\expandafter{\romannumeral1}.
During inference, given incomplete inputs, CodeBrain predicts the scalar-quantised codes representing the complete brain MRI modality set and decodes them, along with the extracted common features, to impute the missing modality.

\subsection{Compact Representation of Brain MRI}
The purpose of Stage \uppercase\expandafter{\romannumeral1} is to learn a compact representation of the full-stack brain MRI $M_{full}$ via reconstruction.
The uniqueness of Stage \uppercase\expandafter{\romannumeral1} is the design of \textbf{bottleneck feature representation}. It combines two complementary components: 
(1) a \emph{scalar-quantised code map} extracted from $M_{full}$ by a posterior encoder $E_{posterior}$, capturing full-stack characteristics, and 
(2) a \emph{common feature map} extracted from any incomplete input $M_{inc}$ by the shared encoder $E_c$, representing modality-agnostic information.

$M_{full}$ is formed by concatenating the images of $N$ modalities (\eg, T1, T2, PD, FLAIR etc). The incomplete input $M_{inc}$ is formed in the same way, but setting the images of missing modalities with zeros. During training, random masking is applied, where $K$ modalities ($0 < K < N$) are set to zero to simulate diverse imputation scenarios.
We encode $M_{full}$ into a low-dimensional latent feature map $F_{full}$ using encoder $E_{posterior}$, and perform element-wise finite scalar quantisation \cite{mentzerfinite} with levels $L$ as:
\begin{equation}
\label{eq_fsq}
\begin{aligned}
& F_{full} = E_{posterior}(M_{full}), \\
& Z_{full,i} = \lfloor L_i/2 \rfloor \times \tanh(F_{full,i}), \quad i = 0,\dots,d-1, \\
& \hat{Z}_{full} = round(Z_{full}), \\
\end{aligned}
\end{equation}
where $Z_{full} \in \mathbb{R}^{d\times h\times w}$ denotes the feature map with $h \times w$ spatial dimension and $d$-dimensional feature, subscript $i$ denotes the $i$-th feature channel. Each element in $Z_{full,i}$ is a scalar value, bounded within $[-\lfloor L_i/2\rfloor, \lfloor L_i/2\rfloor]$. After rounding, each element of the code map $\hat{Z}_{full}$ contains quantised codes representing an image patch in $M_{full}$.
The $i$-th code channel has $L_i$ integers, yielding a total codebook size of $\prod_{i=0}^{d-1} L_i$. 
To enable end-to-end training, we use the straight-through estimator \cite{van2017neural}, expressing $\hat{Z}_{full}$ as $(Z_{full} + sg[\hat{Z}_{full} - Z_{full}])$, where $sg[\cdot]$ is the stop-gradient operation that preserves gradients before and after $round$.
Finite scalar quantization \cite{mentzerfinite} eliminates the need for an explicit learnable codebook and avoids auxiliary regularisation losses \cite{yu2022vectorquantized}, ensuring efficient and stable training.

Since quantisation is inherently lossy and the code space is low-dimensional ($d$ is small), it may not restore all details of $M_{full}$. We incorporate an additional common feature encoder $E_c$ to extract features $F_c$ from input images to enhance the reconstruction quality. Note that $F_c$ is modality-agnostic, which is achieved by extracting $F_c$ from arbitrary combinations of incomplete modalities $M_{inc}$ and enforcing it to contribute to the synthesis of the complete brain MRI modality set. Specifically, we have
\begin{equation}
\label{eq_rec}
\begin{aligned}
F_c = E_{c}(M_{inc}), \ \ 
\Tilde{M}_{full} = D(Concat[\hat{Z}_{full}, F_c]),
\end{aligned}
\end{equation}
where $\Tilde{M}_{full}$ denotes the reconstruction for all modalities from the concatenation of the quantised code map $\hat{Z}_{full}$ with common features $F_c$ by applying the decoder $D$. We train $E_c$, $E_{posterior}$ and $D$ by optimising the following reconstruction loss:
\begin{equation}
\begin{aligned}
\mathcal{L}_{rec} = &
\sum_{i=0}^{N-1} \lambda_{[m,a]} \times
\mathcal{L}_{psnr}(\Tilde{M}_{full,i}, M_{full,i}) \\
&+ \mathcal{L}_{gan}(\Tilde{M}_{full}, M_{full}),
\label{eq_loss_rec}
\end{aligned}
\end{equation}
where $\mathcal{L}_{psnr}$ is an approximate differentiable loss of peak signal-to-noise ratio (PSNR) \cite{chen2022simple}, $\mathcal{L}_{gan}$ is an LSGAN loss \cite{isola2017image,liu2023one} employing an $\mathcal{L}_2$ objective to distinguish real and reconstructed images. 
$\lambda_{[m,a]}$ contains only two weights $\lambda_m$ and $\lambda_a$ to assign different importance to masked and available modalities of $M_{inc}$, as \cite{liu2023one,meng2024multi,zhang2024unified}.

In summary, this design offers three key advantages:
(1) All full-stack brain MRI samples are projected into a shared low-dimensional quantised space, effectively capturing cross-modality relationships and reducing inter-modality discrepancies;
(2) The scalar-quantised codes $\hat{Z}_{full}$ represent region-level modality-specific characteristics, while the common features $F_c$ encode modality-shared information, providing a smooth bridge to Stage~\uppercase\expandafter{\romannumeral2}; and
(3) The model requires only two loss terms for optimisation, avoiding complicated tuning and facilitating stable training.

\subsection{Look-up of Full-stack Quantised Codes}
After Stage \uppercase\expandafter{\romannumeral1} learns full-stack codes in the quantised space, Stage \uppercase\expandafter{\romannumeral2} performs the code look-up for imputation.
Here, a prior encoder $E_{prior}$ is trained to predict full-stack codes $\Tilde{Z}_{full}$ from incomplete data $M_{inc}$, supervised by $\hat{Z}_{full}$ obtained from Stage \uppercase\expandafter{\romannumeral1}:
\begin{equation}
\label{eq_pred}
\begin{aligned}
\Tilde{Z}_{full} = E_{prior}(M_{inc}), \ \ 
\mathcal{L}_{pred} = \mathcal{D}(\Tilde{Z}_{full}, \hat{Z}_{full}),
\end{aligned}
\end{equation}
where $\mathcal{D}(\cdot)$ measures the distance between the predicted code map $\Tilde{Z}_{full}$ and target code map $\hat{Z}_{full}$.
A straightforward approach is to formulate this as a classification problem, where each element of $\hat{Z}_{full}$ corresponds to a quantised integer level, and $\mathcal{D}$ can be implemented as a cross-entropy loss.
However, such a categorical formulation assumes that all quantised codes are independent and equally distant, overlooking the intrinsic \emph{clustering structure} of the quantised code space, where adjacent codes correspond to semantically similar image patches \cite{zheng2023online}.

To better capture this smooth relationship, we consider the prediction task as a \textbf{grading} (or ordinal regression) problem.
Instead of predicting a single discrete label, each scalar value is represented as a sequence of ordered binary decisions that collectively reflect its magnitude.
For example, the $i$-th channel of the code space has $L_i$ discrete levels and the ground-truth label is $y_i \in [0,1,\dots,L_{i-1}]$. We construct its ordinary grading array $o^i$ of size $(L_i-1)$:
\begin{equation}
\label{eq_trans}
\begin{aligned}
& o^i_j = 
\begin{cases} 
1 & \text{if } j < y_i \\
0 & \text{else }
\end{cases}
\end{aligned},
\end{equation}
where $y_i$ can be reverted by computing the sum of $o^i$. By using the above transformation, we then employ a binary cross-entropy loss to train the prior encoder $E_{prior}$:
\begin{equation}
\label{eq_grad}
\mathcal{L}_{grad} = \mathcal{L}_{bce}(\Tilde{O}_{full}, \hat{O}_{full}),
\end{equation}
where $\Tilde{O}_{full}$ and $\hat{O}_{full}$ are the transformed codes of $\Tilde{Z}_{full}$ and $\hat{Z}_{full}$
This formulation explicitly encodes the clustering structure of $Q$, enabling smoother code transitions.
\begin{table*}[htbp]
\caption{\textbf{Quantitative comparisons of CodeBrain in different imputation scenarios of brain MRI on IXI.} The column scenario denotes the available input modalities. Performance is shown as ``\textit{Imputation} (\textit{Reconstruction})''. }
\centering
\resizebox{1.0\linewidth}{!}{
\begin{tabular}{ccc|ccc|ccc|ccc}
\toprule[1.5pt]
\multicolumn{3}{c|}{Scenarios} & \multicolumn{3}{c|}{T1 (Missing)} & \multicolumn{3}{c|}{T2 (Missing)}  & \multicolumn{3}{c}{PD (Missing)} \\
\midrule
T1 & T2 & PD & PSNR (dB) $\uparrow$ & SSIM (\%) $\uparrow$ & MAE ($\times$1000) $\downarrow$ & PSNR (dB) $\uparrow$ & SSIM (\%) $\uparrow$ & MAE ($\times$1000) $\downarrow$ & PSNR (dB) $\uparrow$ & SSIM (\%) $\uparrow$ & MAE ($\times$1000) $\downarrow$
\\ 
\midrule
&&$\surd$ &28.51 (35.45)&93.51 (97.36)&17.95 (8.52) &30.08 (33.09)&93.76 (95.12)&15.92 (12.00)  &\multicolumn{3}{c}{N/A}\\
&$\surd$& &28.08 (35.09)&93.20 (97.15)&19.04 (8.92) &\multicolumn{3}{c|}{N/A} &33.42 (36.87)&96.12 (97.29)&11.59 (8.00)\\
$\surd$ && &\multicolumn{3}{c|}{N/A} &23.61 (27.82)&86.20 (91.23)&30.54 (19.93)&27.10 (31.83)&89.97 (94.32)&21.26 (12.84) \\
\midrule
$\surd$ &$\surd$& &\multicolumn{3}{c|}{N/A} &\multicolumn{3}{c|}{N/A}&34.65 (38.17)&96.48 (97.65)&10.15 (7.02)\\
$\surd$ &&$\surd$ &\multicolumn{3}{c|}{N/A} &31.08 (34.16)&94.16 (95.48)&14.76 (10.97) &\multicolumn{3}{c}{N/A} \\
&$\surd$&$\surd$ &28.95 (36.36)&94.01 (97.85) &16.93 (7.64) &\multicolumn{3}{c|}{N/A}  &\multicolumn{3}{c}{N/A} \\
\midrule
\multicolumn{3}{c|}{$mean$} &28.51 (35.64) &93.57 (97.45) &17.97 (8.36)&28.26 (31.69)&91.37 (93.94)&20.41 (14.30)&31.72 (35.63) &94.19 (96.42)&14.33 (9.29)\\
\bottomrule[1.5pt]
\end{tabular}}
\label{tab_ixi_all}
\end{table*}
\begin{figure*}[htp]
  \centering
    \includegraphics[width=0.9\linewidth]{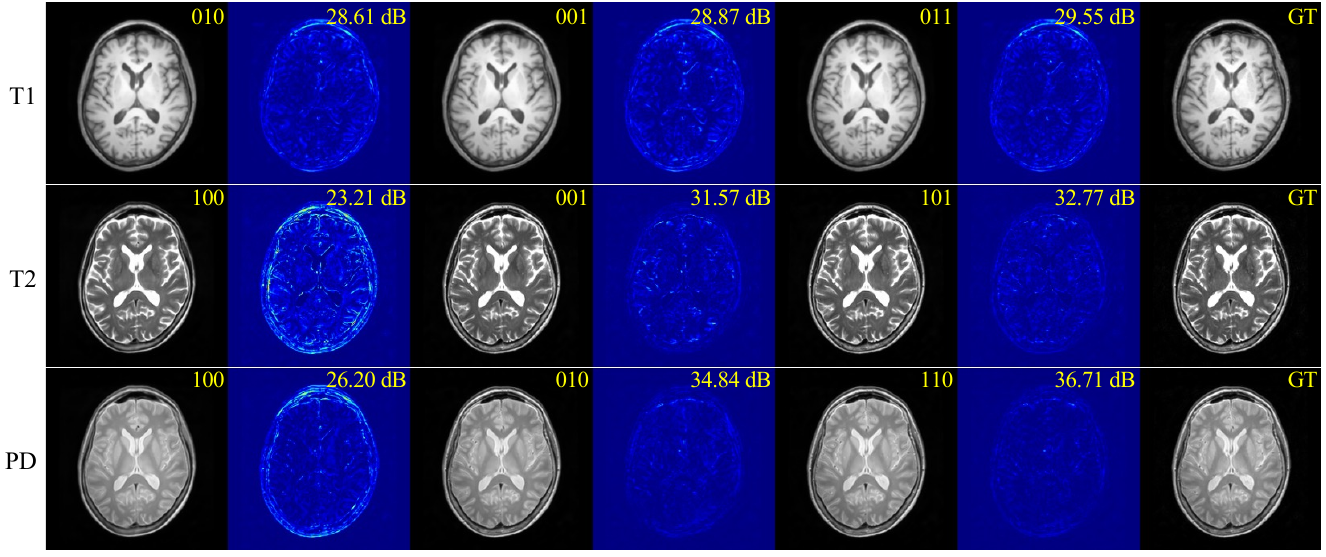}
   \caption{\textbf{Exemplar synthesized brain MRI scans of CodeBrain on IXI}. 1st and 3rd columns: O$\rightarrow$O imputed results; 5th column: M$\rightarrow$O imputed results; 7th column: the original modality. Each synthesized scan is equipped with its corresponding error map on the right. The three bits in an annotation ``010'' correspond to T1, T2, and PD, respectively, where ``1'' indicates that modality is available.}
   \label{fig_results}
\end{figure*}

The advantages of this grading-based design are: 
(1) Cross-modality translation is simplified to predicting full-stack codes rather than intensity values;
(2) The grading-based loss preserves the smooth distributions of the quantised code space, improving prediction stability and precision; and
(3) All modules remain modality-agnostic, removing the need for any modality-specific designs.

\section{Experiments and Results}
\label{sec:experiments}

\subsection{Implementation Details}
\noindent\textbf{Datasets.}
We evaluated our CodeBrain on the IXI\footnote{\url{https://brain-development.org/ixi-dataset/}} and BraTS 2023 \cite{labella2023asnr} datasets. IXI contains non-skull-stripped MRI scans of 577 healthy subjects, which were scanned at three London hospitals with different MRI machines. Each subject contains T1, T2, and proton density-weighted (PD) modalities. Since these modalities are not spatially registered, we use  ANTsPy\footnote{\url{https://github.com/ANTsX/ANTsPy}} for the cross-modal registration (SyNRA). Using the same setting as \cite{liu2023one,meng2024multi}, 90 transverse brain slices are extracted from the middle of each 3D volume. We then crop these slices into a fixed size of 256$\times$256 and randomly select 500 subjects for training, 37 for validation, and the remaining 40 for testing.

BraTS 2023 comprises multi-site multi-parametric MRI (mpMRI) scans of brain tumour patients, including T1, T2, FLAIR, and T1Gd modalities. Each scan is skull-stripped and rigidly registered to the same space. As \cite{liu2023one,meng2024multi}, we extract the middle 80 transverse slices in our experiments, which are further cropped to a fixed size of 240$\times$240. The training, validation, and test sets include 500, 40, and 40 randomly selected subjects, respectively.

\begin{table*}[htbp]
\caption{\textbf{Comparison results on the IXI (Top) and BraTS 2023 (Bottom) datasets}. Here, ``O$\rightarrow$O'' and ``M$\rightarrow$O'' indicate one-to-one and many-to-one scenarios. Our CodeBrain surpasses all other methods of mean performance (Red and Blue indicate the top two methods).}
\centering
\resizebox{0.75\linewidth}{!}{
\begin{tabular}{c|ccc|ccc|ccc}
\toprule[1.5pt]
\multirow{2}{*}{\diagbox{Methods}{Metrics}} & \multicolumn{3}{c|}{PSNR (dB) $\uparrow$}  & \multicolumn{3}{c|}{SSIM (\%) $\uparrow$} & \multicolumn{3}{c}{MAE ($\times$1000) $\downarrow$}\\
\cline{2-10} & $mean$ & O$\rightarrow$O & M$\rightarrow$O  & $mean$ & O$\rightarrow$O & M$\rightarrow$O & $mean$ & O$\rightarrow$O & M$\rightarrow$O
\\ 
\midrule
MMGAN \cite{sharma2019missing} &27.64&26.76&29.41 &90.84&89.71&93.10 &21.39&23.31&17.55 \\
MMT \cite{liu2023one} &28.06&27.11&29.96 &91.42&90.30&93.65&20.21&22.19&16.25 \\
M2DN \cite{meng2024multi} &28.14&27.38&29.67&91.80&90.95&93.49&19.69&21.22&16.64 \\
Zhang et al. \cite{zhang2024unified} &\cellcolor{blue!15}29.00&\cellcolor{blue!15}28.08&\cellcolor{blue!15}30.85&\cellcolor{blue!15}92.63&\cellcolor{blue!15}91.63&\cellcolor{blue!15}94.62&\cellcolor{blue!15}18.40&\cellcolor{blue!15}20.10&\cellcolor{blue!15}15.00 \\
MMHVAE \cite{dorent2025unified} &28.11&27.23&29.88 &91.20&90.15&93.29 &20.14&21.96&16.51 \\
\textbf{Our CodeBrain} &\cellcolor{red!15}\textbf{29.50}&\cellcolor{red!15}\textbf{28.47}&\cellcolor{red!15}\textbf{31.56}
&\cellcolor{red!15}\textbf{93.05}&\cellcolor{red!15}\textbf{92.13}&\cellcolor{red!15}\textbf{94.88} &\cellcolor{red!15}\textbf{17.57}&\cellcolor{red!15}\textbf{19.38}&\cellcolor{red!15}\textbf{13.95} \\
\midrule
MMGAN \cite{sharma2019missing} &24.28&23.76&24.68&89.11&88.22&89.78 &23.72&25.15&22.65 \\
MMT \cite{liu2023one} &24.58&23.88&25.11 &89.47&88.38&90.30 &22.65&24.55&21.22 \\
M2DN \cite{meng2024multi} &24.34&23.69&24.83 &89.65&88.72&90.35 &22.95&24.70&21.64 \\
Zhang et al. \cite{zhang2024unified} &\cellcolor{blue!15}25.01&\cellcolor{blue!15}24.31&\cellcolor{blue!15}25.54&\cellcolor{blue!15}89.98&\cellcolor{blue!15}89.00&\cellcolor{blue!15}90.72 &\cellcolor{blue!15}21.53&\cellcolor{blue!15}23.31&\cellcolor{blue!15}20.20 \\
MMHVAE \cite{dorent2025unified} &24.29&23.66&24.76 &88.83&87.88&89.54 &23.63&25.46&22.25 \\
\textbf{Our CodeBrain} &\cellcolor{red!15}\textbf{25.31}&\cellcolor{red!15}\textbf{24.67}&\cellcolor{red!15}\textbf{25.79} &\cellcolor{red!15}\textbf{90.49}&\cellcolor{red!15}\textbf{89.64}&\cellcolor{red!15}\textbf{91.12}
&\cellcolor{red!15}\textbf{21.01}&\cellcolor{red!15}\textbf{22.57}&\cellcolor{red!15}\textbf{19.84} \\
\bottomrule[1.5pt]
\end{tabular}}
\label{tab_comparison}
\end{table*}
\begin{figure*}[htp]
  \centering
   \includegraphics[width=0.9\linewidth]{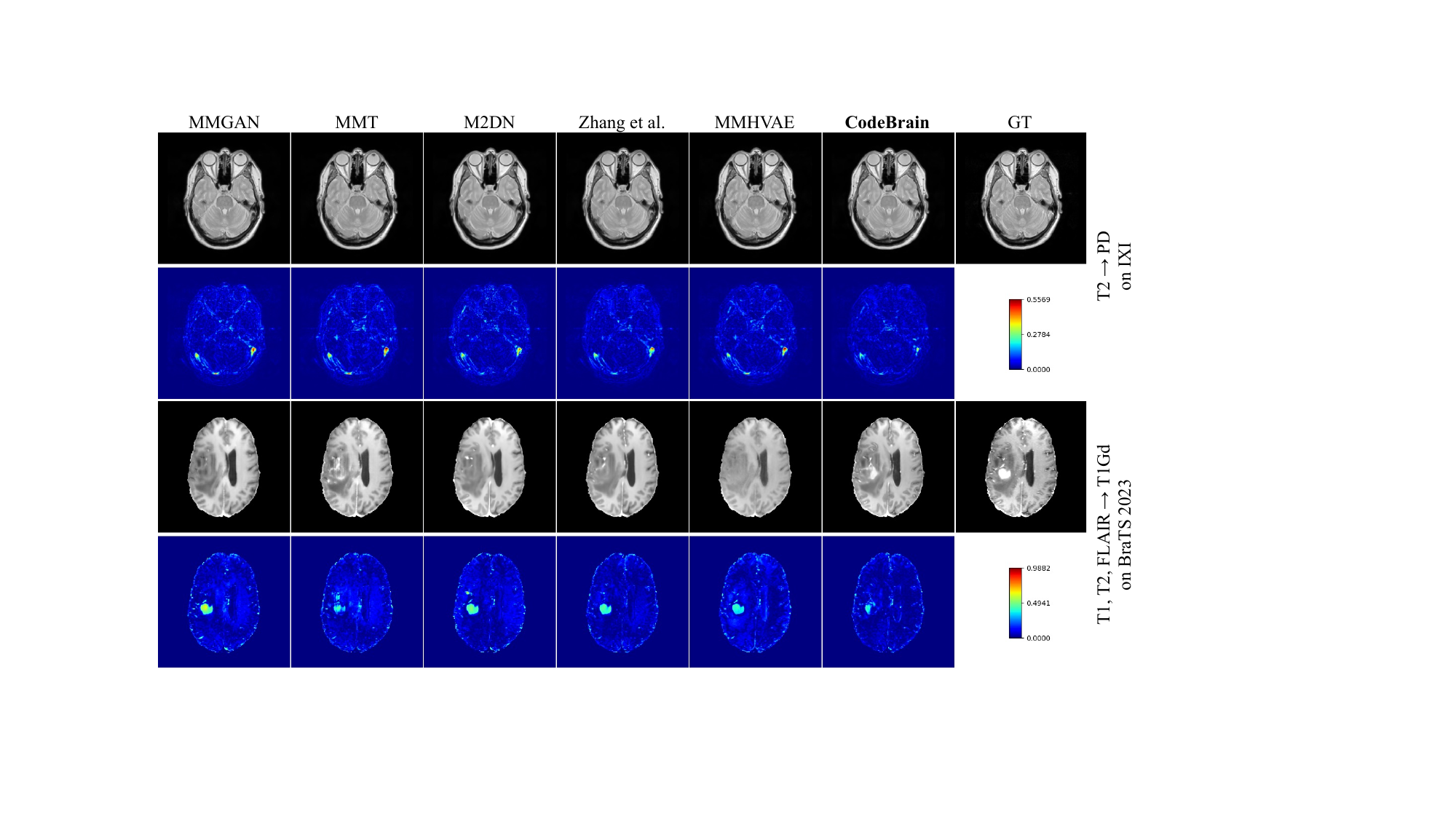}
   \caption{\textbf{Exemplar comparison results on the IXI (Top) and BraTS 2023 (Bottom) datasets}. We show results between five public models and our CodeBrain in the T2$\rightarrow$PD (Top) and T1, T2, FLAIR$\rightarrow$T1Gd (Bottom) scenarios, along with corresponding error maps.}
   \label{fig_compare}
\end{figure*}
\noindent\textbf{Setting.}
We first normalised these slices into a fixed intensity range of 0 to 1 by a min-max normalisation, making the voxel intensities across different subjects and different modalities comparable, same as \cite{zhang2024unified}. For the training loss, we set $\lambda_m$ to 20, $\lambda_a$ to 5, and the batch size to 48. As suggested in \cite{mentzerfinite}, $L$ is set as $[8, 8, 8, 5, 5, 5]$ so that $d$ is 6, indicating the total codebook size of 64,000.
We adopted the NAFNet \cite{chen2022simple} as the backbone and used the AdamW optimizer with an initial learning rate of 1e-4. 
All experiments were conducted in an identical environment for fair comparisons (Hardware: 8$\times$ NVIDIA GeForce 4090 GPUs; Software: PyTorch: 2.8.0, CUDA: 12.4, Random Seed: 1334). Each stage has 300 epochs, totalling 2.38 training days.

\noindent\textbf{Evaluation Metrics.}
We use three metrics to evaluate the performance: PSNR, structural similarity index (SSIM), and mean absolute error (MAE). We compare CodeBrain with five public unified imputation models: MMGAN \cite{sharma2019missing}, transformer-based MMT \cite{liu2023one}, diffusion-based M2DN \cite{meng2024multi}, Zhang et al. \cite{zhang2024unified}, and MMHVAE \cite{dorent2025unified}. We re-implemented the approaches \cite{liu2023one,meng2024multi,zhang2024unified} strictly according to their works. We will also release our settings to establish a public benchmark for the brain MRI imputation.

\subsection{Imputation in Different Scenarios}
\label{sec:unified_results}

Table~\ref{tab_ixi_all} presents the quantitative results for different imputation scenarios on IXI. It shows that PD can be more easily synthesized from other modalities, whereas T2 is challenging to generate from T1, reflecting their distinct examination objectives in clinical practice.
Furthermore, different one-to-one (``O$\rightarrow$O'') imputation results (\ie, the top part of Table~\ref{tab_ixi_all}) show that T1 can be better translated from PD than T2, while T2 and PD are highly related, as stated in \cite{lee2020assessing,liu2023one}.
Then, many-to-one (``M$\rightarrow$O'') settings can achieve better performance in both reconstruction and imputation.

Fig.~\ref{fig_results} further shows visualized results in different scenarios. We can see that the proposed CodeBrain model generates accurate and plausible anatomic structures for different imputation scenarios of brain MRI modalities.

\begin{figure*}[htp]
  \centering
  \includegraphics[width=0.9\linewidth]{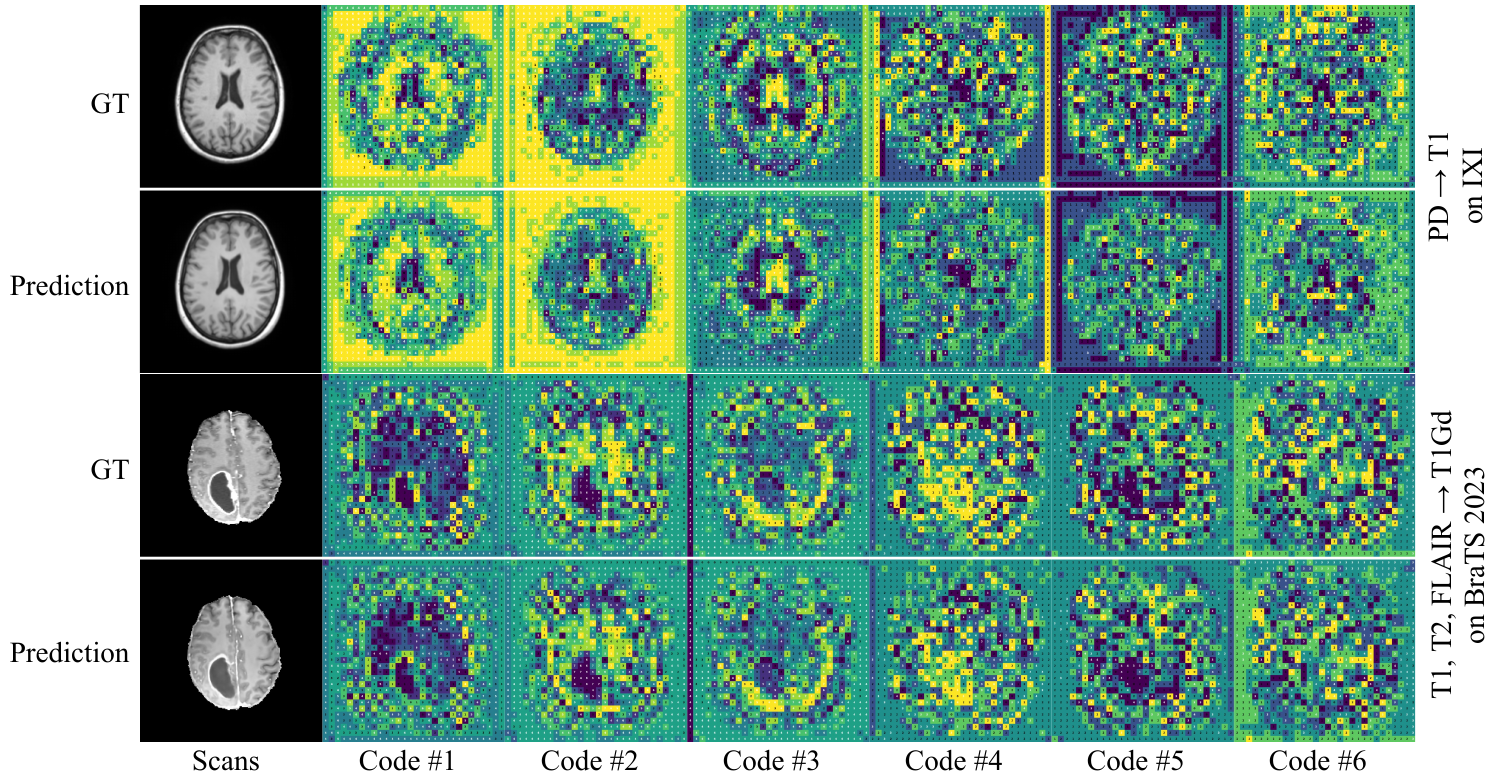}
   \caption{\textbf{Exemplar code maps on the IXI (Top) and BraTS 2023 (Bottom) datasets}. Using $L=[8,8,8,5,5,5]$, we visualize all codes under the ``PD$\rightarrow$T1'' and ``T1, T2, FLAIR$\rightarrow$T1Gd'' scenarios, showing clustered distributions in the latent quantised space $Q$.}
   \label{fig_code}
\end{figure*}

\subsection{Comparisons}
Table~\ref{tab_comparison} compares the results of CodeBrain to five existing models \cite{sharma2019missing,liu2023one,meng2024multi,zhang2024unified,dorent2025unified} for unified brain MRI imputation on the IXI (Top) and BraTS 2023 (Bottom) datasets. It reveals that the CodeBrain substantially outperforms other methods for various ``O$\rightarrow$O'' and ``M$\rightarrow$O'' settings on both datasets. For example, CodeBrain outperforms the second-best work \cite{zhang2024unified} by 0.50 dB in PSNR on IXI.
Furthermore, although training the CodeBrain model only adopts PSNR-related losses without applying any structural loss for supervision, it achieves an average 0.42\% SSIM gain over \cite{zhang2024unified}. This indicates that CodeBrain can achieve invariant anatomical representation among different modalities \cite{labella2023asnr}.

Table~\ref{tab_comparison} (Bottom) further shows that CodeBrain outperforms all other methods by all metrics on BraTS 2023. Even with inconsistent regions (\eg, with or without clear tumour boundaries), the proposed model achieves the best performance, especially in SSIM. This is critical in clinical deployments to maintain a high structural consistency among different brain MRI modalities.
Fig.~\ref{fig_compare} provides visual comparison results on IXI and BraTS 2023, respectively. It can be seen that CodeBrain has fewer synthesis errors than other methods (\eg, the brain tissues and enhanced tumours).
\subsection{Discussions}

\noindent\textbf{Effects of Common Features.} The common feature map $F_c$ is designed to capture modality-agnostic information that can be directly extracted from any incomplete input. This allows $E_{prior}$ to concentrate solely on predicting full-stack characteristics, \ie, quantised codes. 
As shown in Table~\ref{tab_ixi_ablation},  incorporating $F_c$ significantly enhances the reconstruction performance by a 4.17 dB PSNR gain on IXI.

\noindent\textbf{Effects of Grading.} The grading loss~\eqref{eq_grad} enforces the smoothness characteristic of the quantised space $Q$. Table~\ref{tab_ixi_ablation} (Bottom) shows that this design simplifies the code prediction and improves the imputation performance on IXI.

\noindent\textbf{Effects of $\mathcal{L}_{psnr}$.} We follow the standard configuration of our backbone~\cite{chen2022simple} and adopt the PSNR loss for image reconstruction. Using the traditional $\mathcal{L}_1$ loss yields a PSNR of 34.08 dB and an SSIM of 96.28\%. In contrast, $\mathcal{L}_{psnr}$ improves the PSNR to 34.32 dB, albeit with a slight reduction in SSIM to 95.94\%. Future work will investigate more losses \cite{tomaszewski2021biological} to further enhance reconstruction quality.

\noindent\textbf{Brain Tumour Segmentation.} 
Following \cite{labella2023asnr}, we perform brain tumour segmentation to show the added value of imputation, using a 3D U-Net model trained on BraTS 2023. 3D Dice scores are reported in Table~\ref{tab_seg} across four ``$3\rightarrow1$" imputation scenarios.
The results reveal that:
(1) Simply imputing missing modalities with zero values leads to a significant performance drop (see the 1st row). For example, missing the FLAIR modality will lead to the failed segmentation directly.
(2) Leveraging synthesised modalities for segmentation with CodeBrain outperforms other imputation methods. Especially, our model works well in imputing contrast-enhanced modalities \ie, T1Gd, without any particular design.
(3) CodeBrain’s imputed modalities achieve segmentation performance comparable to the upper bound, \ie, training a segmentation model using all real brain MRI modalities (see the 5th row vs. 6th row).

\noindent\textbf{Selection of $\lambda_m$ and $\lambda_a$.}
Similar to~\cite{liu2023one,zhang2024unified,meng2024multi}, we assign different weights to reconstruct the available and masked modalities of $M_{inc}$. Table~\ref{tab_lamda} shows that setting $\lambda_m$ and $\lambda_a$ to 20 and 5 yields the best performance on IXI.

\noindent\textbf{Quantised Codes.}
To visualize the distributions of quantised codes in the code space, we further show code maps of CodeBrain under the ``PD$\rightarrow$T1'' (Top) and ``T1, T2, FLAIR$\rightarrow$T1Gd'' (Bottom) scenarios in Fig.~\ref{fig_code}. We can see that, without any training regularisation, the code distributions exhibit clustering characteristics. These clustered codes reflect coarse anatomical structures of the brain, potentially bridging image synthesis and perception tasks (\eg, brain segmentation \cite{zhang2021overview}). Furthermore, Stage \uppercase\expandafter{\romannumeral2} accurately predicts most of the corresponding codes and synthesises high-quality brain MRI modalities on both datasets.

\begin{figure}[b]
  \centering
   \includegraphics[width=0.98\linewidth]{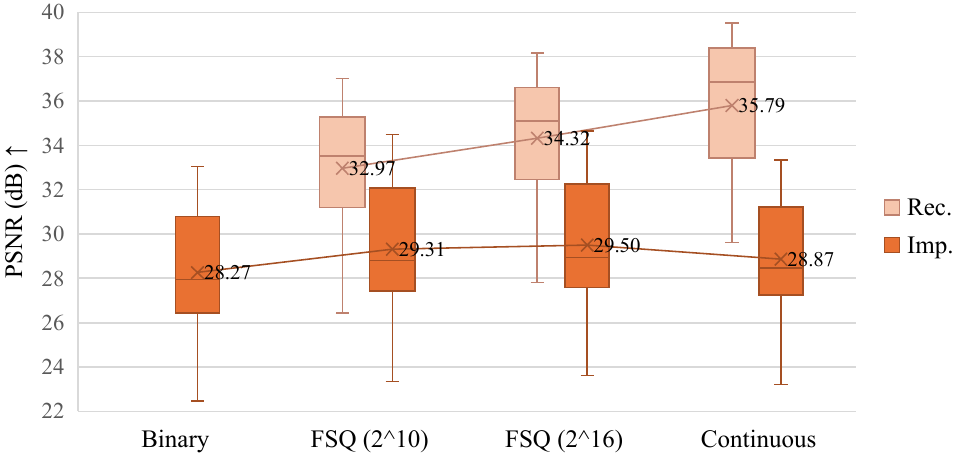}
   \caption{\textbf{Comparisons of different conditions on IXI}. Compared with using fixed binary conditions (1st column) and infinite continuous conditions (4th column), the discrete codes (2nd and 3rd columns) achieve higher PSNR performance.}
   \label{fig_size}
\end{figure}

\noindent\textbf{Selection of Conditions.}
Fig.~\ref{fig_size} compares four different condition designs. Compared with fixed binary conditions, the proposed learnable region-level conditions achieve superior PSNR performance for both reconstruction and imputation on IXI, indicating their ability to capture MRI characteristics. However, when the condition space becomes overly complex (\ie, infinite continuous variables, see the 4th column of Fig.~\ref{fig_size}), the imputation task becomes substantially more difficult, leading to degraded performance. In contrast, quantised code representations provide a favourable balance between expressiveness and tractability. Therefore, following~\cite{mentzerfinite}, we set the quantisation level to $L=[8,8,8,5,5,5]$, approximately corresponding to $2^{16}$ potential scalar variables in FSQ \cite{mentzerfinite}.

\noindent\textbf{Limitation and Future Work.} Fig.~\ref{fig_compare} shows that despite achieving the superior imputation performance, our model may still be impacted by hallucination. Further improvements could come from 3D modelling \cite{guo2025maisi}, and the incorporation of MRI physical theories \cite{plewes2012physics}, particularly for contrast-enhanced modalities such as T1Gd.

\begin{table}[htbp]
\caption{\textbf{Ablation studies of our CodeBrain on IXI}. For reconstruction (Rec.), we compare different training settings, including models trained without and with $F_c$. For imputation (Imp.), we further evaluate two prediction designs: classification-based (Cls.) and our proposed grading-based scheme (Grad.). Here, we report the mean performance across nine imputation scenarios.}
\centering
\resizebox{0.85\linewidth}{!}{
\begin{tabular}{c|c|ccc}
\toprule[1.5pt]
Stages&Settings& PSNR (dB) $\uparrow$ & SSIM (\%) $\uparrow$ & MAE ($\times$1000) $\downarrow$\\
\midrule
\multirow{2}{*}{Rec.}&w/o $F_c$&30.15&92.75&15.87\\
% \midrule
% \multirow{2}{*}{Rec.}& w/ L1&34.08&96.28&10.43 \\
& w/ $F_c$&\cellcolor{red!15}\textbf{34.32}&\cellcolor{red!15}\textbf{95.94}&\cellcolor{red!15}\textbf{10.65} \\
\midrule
\multirow{2}{*}{Imp.}& w/ Cls.&29.24&92.83&18.00\\
& w/ Grad.&\cellcolor{red!15}\textbf{29.50}&\cellcolor{red!15}\textbf{93.05}&\cellcolor{red!15}\textbf{17.57} \\
\bottomrule[1.5pt]
\end{tabular}}
\label{tab_ixi_ablation}
\end{table}

\begin{table}[htbp]
\caption{\textbf{Comparisons of brain tumour segmentation across four ``3$\rightarrow$1'' scenarios on BraTS 2023}. Results are shown for zero imputation (1st row), synthesized imputation (2nd–5th rows), and full-modality input (6th row). A 3D U-Net is trained for segmentation, and the mean scores of 3D Dice (\%, $\uparrow$) are reported for the enhancing tumour, tumour core, and whole tumour.
}
\centering
\resizebox{0.8\linewidth}{!}{
\begin{tabular}{c|cccc|c}
\toprule[1.5pt]

&T1&T2&FLAIR&T1Gd&$mean$ \\

\midrule
Zero Imputation &47.58&77.39&0.84&27.75&38.39\\
\midrule
MMT \cite{liu2023one} &86.07&\cellcolor{red!15}\textbf{87.08}&85.90&\cellcolor{blue!15}57.25&\cellcolor{blue!15}79.08\\
Zhang et al. \cite{zhang2024unified} &\cellcolor{red!15}\textbf{86.26}&85.90&86.31&47.35&76.46\\
MMHVAE \cite{dorent2025unified} &85.90&85.02&\cellcolor{red!15}\textbf{86.73}&40.22&74.47\\
\textbf{Our CodeBrain} &\cellcolor{blue!15}86.10&\cellcolor{blue!15}86.23&\cellcolor{blue!15}86.47&\cellcolor{red!15}\textbf{61.44}&\cellcolor{red!15}\textbf{80.06}\\
\midrule
Full Modalities &\multicolumn{4}{c}{87.40}\\
\bottomrule[1.5pt]
\end{tabular}}
\label{tab_seg}
\end{table}

\section{Conclusion}
\label{sec:conclusion}
We have presented CodeBrain, a unified model for virtual full-stack scanning of brain MRI. The key idea is to cast the cross-modality translation task into two stages: learning compact codes and looking up these codes. First, each complete brain MRI set is compressed to a quantised code map using finite scalar quantisation. Second, CodeBrain predicts the full-stack code map and concatenates it with common features for the target synthesis. Extensive experiments demonstrated its superior imputation performance.

\begin{table}[h]
\caption{\textbf{Ablation studies of different loss weights on IXI}.}
\label{tab_lamda}
\centering
\begin{subtable}{0.35 \linewidth}
\centering
\resizebox{\linewidth}{!}
{
\begin{tabular}{c|c|c}
\toprule[1.5pt]
$\lambda_m$  & $\lambda_a$  & PSNR (dB) $\uparrow$ \\ 
\midrule
10 & \multirow{3}{*}{\, 5 \,}  & 33.83   \\
20 &  &\cellcolor{red!15}\textbf{34.32}   \\
30 &   &34.21    \\
\bottomrule[1.5pt]
\end{tabular}
}
\end{subtable}%
\begin{subtable}{0.36\linewidth}
\centering
\resizebox{\linewidth}{!}
{
\begin{tabular}{c|c|c}
\toprule[1.5pt]
$\lambda_m$  & $\lambda_a$  & PSNR (dB) $\uparrow$ \\ 
\midrule
\multirow{3}{*}{\, 20 \,} & 3  &  34.26  \\
 & 5 &\cellcolor{red!15}\textbf{34.32}   \\
&   7&   34.01 \\
\bottomrule[1.5pt]
\end{tabular}
}
\end{subtable}%
\end{table}

\section{Acknowledgement}
Y. Wu and W. Bai acknowledge the support of the EPSRC CVD-Net Programme Grant (EP/Z531297/1) and NIHR Imperial Biomedical Research Centre (BRC). The views expressed are those of the authors and not necessarily those of the NIHR or the Department of Health and Social Care. This research was also partially supported by the Department of Health, Disability and Ageing (Australia) under Grant NCRI000074.

{
    \small
    \bibliographystyle{ieeenat_fullname}
    \bibliography{main}

@String(PAMI = {IEEE Trans. Pattern Anal. Mach. Intell.})

@String(IJCV = {Int. J. Comput. Vis.})

@String(CVPR= {IEEE Conf. Comput. Vis. Pattern Recog.})

@String(ICCV= {Int. Conf. Comput. Vis.})

@String(ECCV= {Eur. Conf. Comput. Vis.})

@String(NIPS= {Adv. Neural Inform. Process. Syst.})

@String(ACMMM= {ACM Int. Conf. Multimedia})

@String(ICLR = {Int. Conf. Learn. Represent.})

@String(PAMI  = {IEEE TPAMI})

@String(IJCV  = {IJCV})

@String(CVPR  = {CVPR})

@String(ICCV  = {ICCV})

@String(ECCV  = {ECCV})

@String(NIPS  = {NeurIPS})

@String(ACMMM = {ACM MM})

@String(ICLR  = {ICLR})

@String(MICCAI = {Int. Conf. Medical Image Computing and Computer-Assisted Intervention})

@String(MICCAI = {MICCAI})

@String(WACV = {Winter Conf. Appl. Comput. Vis.})

@String(WACV = {WACV})

@String(TMI = {{IEEE Trans. Medical Imaging}})

@String(TMI = {{IEEE TMI}})

@article{xie2023cross,
  title={{Cross-modality neuroimage synthesis: A survey}},
  author={Xie, Guoyang and Huang, Yawen and Wang, Jinbao and Lyu, Jiayi and Zheng, Feng and Zheng, Yefeng and Jin, Yaochu},
  journal={ACM Computing Surveys},
  volume={56},
  number={3},
  year={2023}
}

@inproceedings{zhao2025reverse,
  title={{Reverse Imaging for Wide-Spectrum Generalization of Cardiac MRI Segmentation}},
  author={Zhao, Yidong and Kellman, Peter and Xue, Hui and Yang, Tongyun and Zhang, Yi and Han, Yuchi and Simonetti, Orlando and Tao, Qian},
  booktitle=MICCAI,
  pages={555--565},
  year={2025}
}

@article{wang2023quantitative,
  title={{Quantitative cerebral blood volume image synthesis from standard MRI using image-to-image translation for brain tumors}},
  author={Wang, Bao and Pan, Yongsheng and Xu, Shangchen and Zhang, Yi and Ming, Yang and Chen, Ligang and Liu, Xuejun and Wang, Chengwei and Liu, Yingchao and Xia, Yong},
  journal={Radiology},
  volume={308},
  number={2},
  year={2023},
  publisher={Radiological Society of North America}
}

@inproceedings{pan2018synthesizing,
  title={{Synthesizing missing PET from MRI with cycle-consistent generative adversarial networks for Alzheimer’s disease diagnosis}},
  author={Pan, Yongsheng and Liu, Mingxia and Lian, Chunfeng and Zhou, Tao and Xia, Yong and Shen, Dinggang},
  booktitle=MICCAI,
  pages={455--463},
  year={2018}
}

@article{liu2023one,
  title={{One model to synthesize them all: Multi-contrast multi-scale transformer for missing data imputation}},
  author={Liu, Jiang and Pasumarthi, Srivathsa and Duffy, Ben and Gong, Enhao and Datta, Keshav and Zaharchuk, Greg},
  journal=TMI,
  volume={42},
  number={9},
  year={2023},
  publisher={IEEE}
}

@article{meng2024multi,
  title={{Multi-modal modality-masked diffusion network for brain MRI synthesis with random modality missing}},
  author={Meng, Xiangxi and Sun, Kaicong and Xu, Jun and He, Xuming and Shen, Dinggang},
  journal=TMI,
  volume={43},
  number={7},
  year={2024},
  publisher={IEEE}
}

@article{zhang2024unified,
  title={Unified multi-modal image synthesis for missing modality imputation},
  author={Zhang, Yue and Peng, Chengtao and Wang, Qiuli and Song, Dan and Li, Kaiyan and Zhou, S Kevin},
  journal=TMI,
  year={2024},
  publisher={IEEE}
}

@article{billot2023synthseg,
  title={{SynthSeg: Segmentation of brain MRI scans of any contrast and resolution without retraining}},
  author={Billot, Benjamin and Greve, Douglas N and Puonti, Oula and Thielscher, Axel and Van Leemput, Koen and Fischl, Bruce and Dalca, Adrian V and Iglesias, Juan Eugenio and others},
  journal={Medical Image Analysis},
  volume={86},
  year={2023},
  publisher={Elsevier}
}

@article{dalmaz2022resvit,
  title={{ResViT: residual vision transformers for multimodal medical image synthesis}},
  author={Dalmaz, Onat and Yurt, Mahmut and {\c{C}}ukur, Tolga},
  journal=TMI,
  volume={41},
  number={10},
  year={2022},
  publisher={IEEE}
}

@article{dar2019image,
  title={{Image synthesis in multi-contrast MRI with conditional generative adversarial networks}},
  author={Dar, Salman UH and Yurt, Mahmut and Karacan, Levent and Erdem, Aykut and Erdem, Erkut and Cukur, Tolga},
  journal=TMI,
  volume={38},
  number={10},
  year={2019},
  publisher={IEEE}
}

@article{sharma2019missing,
  title={{Missing MRI pulse sequence synthesis using multi-modal generative adversarial network}},
  author={Sharma, Anmol and Hamarneh, Ghassan},
  journal=TMI,
  volume={39},
  number={4},
  year={2019},
  publisher={IEEE}
}

@article{dalmaz2024one,
  title={{One model to unite them all: Personalized federated learning of multi-contrast MRI synthesis}},
  author={Dalmaz, Onat and Mirza, Muhammad U and Elmas, Gokberk and Ozbey, Muzaffer and Dar, Salman UH and Ceyani, Emir and Oguz, Kader K and Avestimehr, Salman and {\c{C}}ukur, Tolga},
  journal={Medical Image Analysis},
  volume={94},
  year={2024},
  publisher={Elsevier}
}

@article{lee2020assessing,
  title={Assessing the importance of magnetic resonance contrasts using collaborative generative adversarial networks},
  author={Lee, Dongwook and Moon, Won-Jin and Ye, Jong Chul},
  journal={Nature Machine Intelligence},
  volume={2},
  number={1},
  year={2020},
  publisher={Nature Publishing Group UK London}
}

@inproceedings{esser2021taming,
  title={Taming transformers for high-resolution image synthesis},
  author={Esser, Patrick and Rombach, Robin and Ommer, Bjorn},
  booktitle=CVPR,
  pages={12873--12883},
  year={2021}
}

@inproceedings{lee2022autoregressive,
  title={Autoregressive image generation using residual quantization},
  author={Lee, Doyup and Kim, Chiheon and Kim, Saehoon and Cho, Minsu and Han, Wook-Shin},
  booktitle=CVPR,
  pages={11523--11532},
  year={2022}
}

@article{dayarathna2023deep,
  title={{Deep learning based synthesis of MRI, CT and PET: Review and analysis}},
  author={Dayarathna, Sanuwani and Islam, Kh Tohidul and Uribe, Sergio and Yang, Guang and Hayat, Munawar and Chen, Zhaolin},
  journal={Medical Image Analysis},
  volume={92},
  year={2023},
  publisher={Elsevier}
}

@inproceedings{chen2022simple,
  title={Simple baselines for image restoration},
  author={Chen, Liangyu and Chu, Xiaojie and Zhang, Xiangyu and Sun, Jian},
  booktitle=ECCV,
  pages={17--33},
  year={2022}
}

@inproceedings{mentzerfinite,
  title={{Finite scalar quantization: VQ-VAE made simple}},
  author={Mentzer, Fabian and Minnen, David and Agustsson, Eirikur and Tschannen, Michael},
  booktitle=ICLR,
  year = {2024}
}

@inproceedings{williams2020hierarchical,
  title={Hierarchical quantized autoencoders},
  author={Williams, Will and Ringer, Sam and Ash, Tom and MacLeod, David and Dougherty, Jamie and Hughes, John},
  booktitle=NIPS,
  volume={33},
  pages={4524--4535},
  year={2020}
}

@inproceedings{yu2024image,
  title={An Image is Worth 32 Tokens for Reconstruction and Generation},
  author={Yu, Qihang and Weber, Mark and Deng, Xueqing and Shen, Xiaohui and Cremers, Daniel and Chen, Liang-Chieh},
  booktitle=NIPS,
  volume={37},
  pages={128940--128966},
  year={2024}
}

@inproceedings{zheng2022movq,
  title={{MoVQ: Modulating quantized vectors for high-fidelity image generation}},
  author={Zheng, Chuanxia and Vuong, Tung-Long and Cai, Jianfei and Phung, Dinh},
  booktitle=NIPS,
  volume={35},
  pages={23412--23425},
  year={2022}
}

@inproceedings{zheng2023online,
  title={Online clustered codebook},
  author={Zheng, Chuanxia and Vedaldi, Andrea},
  booktitle=ICCV,
  pages={22798--22807},
  year={2023}
}

@inproceedings{tian2024visual,
  title={{Visual autoregressive modeling: Scalable image generation via next-scale prediction}},
  author={Tian, Keyu and Jiang, Yi and Yuan, Zehuan and Peng, Bingyue and Wang, Liwei},
  booktitle=NIPS,
  volume={37},
  pages={84839--84865},
  year={2024}
}

@inproceedings{huang2023towards,
  title={Towards accurate image coding: Improved autoregressive image generation with dynamic vector quantization},
  author={Huang, Mengqi and Mao, Zhendong and Chen, Zhuowei and Zhang, Yongdong},
  booktitle=CVPR,
  pages={22596--22605},
  year={2023}
}

@inproceedings{van2017neural,
  title={Neural discrete representation learning},
  author={Van Den Oord, Aaron and Vinyals, Oriol and others},
  booktitle=NIPS,
  volume={30},
  year={2017}
}

@inproceedings{razavi2019generating,
  title={{Generating diverse high-fidelity images with VQ-VAE-2}},
  author={Razavi, Ali and Van den Oord, Aaron and Vinyals, Oriol},
  booktitle=NIPS,
  volume={32},
  year={2019}
}

@inproceedings{gu2022vector,
  title={Vector quantized diffusion model for text-to-image synthesis},
  author={Gu, Shuyang and Chen, Dong and Bao, Jianmin and Wen, Fang and Zhang, Bo and Chen, Dongdong and Yuan, Lu and Guo, Baining},
  booktitle=CVPR,
  pages={10696--10706},
  year={2022}
}

@inproceedings{tang2024hart,
  title={{HART: Efficient visual generation with hybrid autoregressive transformer}},
  author={Tang, Haotian and Wu, Yecheng and Yang, Shang and Xie, Enze and Chen, Junsong and Chen, Junyu and Zhang, Zhuoyang and Cai, Han and Lu, Yao and Han, Song},
  booktitle=ICLR,
  year={2025}
}

@inproceedings{rombach2022high,
  title={High-resolution image synthesis with latent diffusion models},
  author={Rombach, Robin and Blattmann, Andreas and Lorenz, Dominik and Esser, Patrick and Ommer, Bj{\"o}rn},
  booktitle=CVPR,
  pages={10684--10695},
  year={2022}
}

@inproceedings{hao2024bigr,
  title={{BiGR: Harnessing binary latent codes for image generation and improved visual representation capabilities}},
  author={Hao, Shaozhe and Liu, Xuantong and Qi, Xianbiao and Zhao, Shihao and Zi, Bojia and Xiao, Rong and Han, Kai and Wong, Kwan-Yee K},
  booktitle=ICLR,
  year={2025}
}

@article{ozbey2023unsupervised,
  title={Unsupervised medical image translation with adversarial diffusion models},
  author={{\"O}zbey, Muzaffer and Dalmaz, Onat and Dar, Salman UH and Bedel, Hasan A and {\"O}zturk, {\c{S}}aban and G{\"u}ng{\"o}r, Alper and {\c{C}}ukur, Tolga},
  journal=TMI,
  volume={42},
  number={12},
  pages={3524--3539},
  year={2023},
  publisher={IEEE}
}

@article{gungor2023adaptive,
  title={{Adaptive diffusion priors for accelerated MRI reconstruction}},
  author={G{\"u}ng{\"o}r, Alper and Dar, Salman UH and {\"O}zt{\"u}rk, {\c{S}}aban and Korkmaz, Yilmaz and Bedel, Hasan A and Elmas, Gokberk and Ozbey, Muzaffer and {\c{C}}ukur, Tolga},
  journal={Medical Image Analysis},
  volume={88},
  pages={102872},
  year={2023},
  publisher={Elsevier}
}

@article{tomaszewski2021biological,
  title={The biological meaning of radiomic features},
  author={Tomaszewski, Michal R and Gillies, Robert J},
  journal={Radiology},
  volume={298},
  number={3},
  pages={505--516},
  year={2021},
  publisher={Radiological Society of North America}
}

@article{zhang2021overview,
  title={{Overview of multi-modal brain tumor MR image segmentation}},
  author={Zhang, Wenyin and Wu, Yong and Yang, Bo and Hu, Shunbo and Wu, Liang and Dhelim, Sahraoui},
  journal={Healthcare},
  volume={9},
  number={8},
  pages={1051},
  year={2021},
  organization={MDPI}
}

@inproceedings{he2021checkerboard,
  title={Checkerboard context model for efficient learned image compression},
  author={He, Dailan and Zheng, Yaoyan and Sun, Baocheng and Wang, Yan and Qin, Hongwei},
  booktitle=CVPR,
  pages={14771--14780},
  year={2021}
}

@inproceedings{jang2016categorical,
  title={{Categorical reparameterization with Gumbel-softmax}},
  author={Jang, Eric and Gu, Shixiang and Poole, Ben},
  booktitle=ICLR,
  year={2017}
}

@article{balakrishnan2019voxelmorph,
  title={{VoxelMorph: a learning framework for deformable medical image registration}},
  author={Balakrishnan, Guha and Zhao, Amy and Sabuncu, Mert R and Guttag, John and Dalca, Adrian V},
  journal=TMI,
  volume={38},
  number={8},
  pages={1788--1800},
  year={2019},
  publisher={IEEE}
}

@article{gilmore2012longitudinal,
  title={Longitudinal development of cortical and subcortical gray matter from birth to 2 years},
  author={Gilmore, John H and Shi, Feng and Woolson, Sandra L and Knickmeyer, Rebecca C and Short, Sarah J and Lin, Weili and Zhu, Hongtu and Hamer, Robert M and Styner, Martin and Shen, Dinggang},
  journal={Cerebral Cortex},
  volume={22},
  number={11},
  pages={2478--2485},
  year={2012},
  publisher={Oxford University Press}
}

@inproceedings{yu2022vectorquantized,
    title={{Vector-quantized image modeling with improved VQGAN}},
    author={Jiahui Yu and Xin Li and Jing Yu Koh and Han Zhang and Ruoming Pang and James Qin and Alexander Ku and Yuanzhong Xu and Jason Baldridge and Yonghui Wu},
    booktitle=ICLR,
    year={2022}
}

@inproceedings{isola2017image,
  title={Image-to-image translation with conditional adversarial networks},
  author={Isola, Phillip and Zhu, Jun-Yan and Zhou, Tinghui and Efros, Alexei A},
  booktitle=CVPR,
  pages={1125--1134},
  year={2017}
}

@article{cao2023autoencoder,
  title={Autoencoder-driven multimodal collaborative learning for medical image synthesis},
  author={Cao, Bing and Bi, Zhiwei and Hu, Qinghua and Zhang, Han and Wang, Nannan and Gao, Xinbo and Shen, Dinggang},
  journal=IJCV,
  volume={131},
  number={8},
  pages={1995--2014},
  year={2023},
  publisher={Springer}
}

@article{labella2023asnr,
  title={{The ASNR-MICCAI brain tumor segmentation (BraTS) challenge 2023: Intracranial meningioma}},
  author={LaBella, Dominic and Adewole, Maruf and Alonso-Basanta, Michelle and Altes, Talissa and Anwar, Syed Muhammad and Baid, Ujjwal and Bergquist, Timothy and Bhalerao, Radhika and Chen, Sully and Chung, Verena and others},
  journal={arXiv preprint arXiv:2305.07642},
  year={2023}
}

@article{plewes2012physics,
  title={{Physics of MRI: a primer}},
  author={Plewes, Donald B and Kucharczyk, Walter},
  journal={Journal of Magnetic Resonance Imaging},
  volume={35},
  number={5},
  pages={1038--1054},
  year={2012},
  publisher={Wiley Online Library}
}

@article{arslan2025self,
  title={Self-consistent recursive diffusion bridge for medical image translation},
  author={Arslan, Fuat and Kabas, Bilal and Dalmaz, Onat and Ozbey, Muzaffer and {\c{C}}ukur, Tolga},
  journal={Medical Image Analysis},
  volume={106},
  pages={103747},
  year={2025},
  publisher={Elsevier}
}

@article{dorent2025unified,
  title={{Unified Cross-Modal Medical Image Synthesis with Hierarchical Mixture of Product-of-Experts}},
  author={Dorent, Reuben and Haouchine, Nazim and Golby, Alexandra and Frisken, Sarah and Kapur, Tina and Wells, William},
  journal=PAMI,
  year={2025},
  publisher={IEEE}
}

@article{tudosiu2024realistic,
  title={Realistic morphology-preserving generative modelling of the brain},
  author={Tudosiu, Petru-Daniel and Pinaya, Walter HL and Ferreira Da Costa, Pedro and Dafflon, Jessica and Patel, Ashay and Borges, Pedro and Fernandez, Virginia and Graham, Mark S and Gray, Robert J and Nachev, Parashkev and others},
  journal={Nature Machine Intelligence},
  volume={6},
  number={7},
  pages={811--819},
  year={2024}
}

@inproceedings{han2025infinity,
  title={{Infinity: Scaling bitwise autoregressive modeling for high-resolution image synthesis}},
  author={Han, Jian and Liu, Jinlai and Jiang, Yi and Yan, Bin and Zhang, Yuqi and Yuan, Zehuan and Peng, Bingyue and Liu, Xiaobing},
  booktitle=CVPR,
  pages={15733--15744},
  year={2025}
}

@inproceedings{zhaoimage,
  title={Image and Video Tokenization with Binary Spherical Quantization},
  author={Zhao, Yue and Xiong, Yuanjun and Kraehenbuehl, Philipp},
  booktitle=ICLR,
  year={2025}
}

@inproceedings{wu2024dataset,
  title={{Dataset, challenge, and evaluation for tumor segmentation variability}},
  author={Wu, Yicheng and Xie, Yutong and Luo, Xiangde and Wu, Qi and Cai, Jianfei},
  booktitle=ACMMM,
  pages={11302--11303},
  year={2024}
}

@inproceedings{zhu2025addressing,
  title={{Addressing representation collapse in vector quantized models with one linear layer}},
  author={Zhu, Yongxin and Li, Bocheng and Xin, Yifei and Xia, Zhihua and Xu, Linli},
  booktitle=ICCV,
  pages={22968--22977},
  year={2025}
}

@inproceedings{hsu2023disentanglement,
  title={{Disentanglement via latent quantization}},
  author={Hsu, Kyle and Dorrell, William and Whittington, James and Wu, Jiajun and Finn, Chelsea},
  booktitle=NIPS,
  volume={36},
  pages={45463--45488},
  year={2023}
}

@inproceedings{guo2025maisi,
  title={Maisi: Medical ai for synthetic imaging},
  author={Guo, Pengfei and Zhao, Can and Yang, Dong and Xu, Ziyue and Nath, Vishwesh and Tang, Yucheng and Simon, Benjamin and Belue, Mason and Harmon, Stephanie and Turkbey, Baris and others},
  booktitle=WACV,
  pages={4430--4441},
  year={2025}
}

@inproceedings{zhang2025structure,
  title={Structure-Aware MRI Translation: Multi-modal Latent Diffusion Model with Arbitrary Missing Modalities},
  author={Zhang, Xinzhe and Liang, Junjie and Cao, Peng and Yang, Jinzhu and Zaiane, Osmar R},
  booktitle=MICCAI,
  year={2025}
}

@article{song2026learning,
  title={Learning Modality-Aware Representations: Adaptive Group-wise Interaction Network for Multimodal MRI Synthesis},
  author={Song, Tao and Wu, Yicheng and Hu, Minhao and Luo, Xiangde and Wei, Linda and Wang, Guotai and Guo, Yi and Xu, Feng and Zhang, Shaoting},
  journal=TMI,
  year={2026},
  publisher={IEEE}
}
}

\end{document}